\documentclass[runningheads]{llncs}
\usepackage{graphicx}

\usepackage{tikz}
\usepackage{comment}
\usepackage{amsmath,amssymb} 
\usepackage{color}
\usepackage{url}
\usepackage[accsupp]{axessibility}  

\begin{document}
\pagestyle{headings}
\mainmatter
\def\ECCVSubNumber{8}  

\title{Using Whole Slide Image Representations from Self-Supervised Contrastive Learning for Melanoma Concordance Regression \thanks{Acknowledgments: The authors thank the support of Jeff Baatz, Ramachandra V. Chamarthi, Nathan Langlois, and Liren Zhu at Proscia for their engineering support; Theresa Feeser, Pratik Patel, and Aysegul Ergin Sutcu at Proscia for their data acquisition and Q\&A support; and Dr. Curtis Thompson at CTA and Dr. David Terrano at Bethesda Dermatology Laboratory for their consensus annotation support.}} 

\titlerunning{Melanoma Concordance Regression}
%

\author{Sean Grullon\inst{1}
\and
Vaughn Spurrier\inst{1}
\and
Jiayi Zhao\inst{1}
\and
Corey Chivers\inst{1}
\and
Yang Jiang\inst{1}
\and
Kiran Motaparthi\inst{2}
\and
Jason Lee \inst{3}
\and
Michael Bonham\inst{1}
\and
Julianna Ianni\inst{1}}

\authorrunning{S. Grullon et al.}

\institute{Proscia, Inc. Philadelphia, United States \and Department of Dermatology, University of Florida College of Medicine \and Department of Dermatology, Sidney Kimmel Medical College at Thomas Jefferson University}
%

\maketitle

\begin{abstract}
Although melanoma occurs more rarely than several other skin cancers, patients' long term survival rate is extremely low if the diagnosis is missed. Diagnosis is complicated by a high discordance rate among pathologists when distinguishing between melanoma and benign melanocytic lesions.  A tool that provides potential concordance information to healthcare providers could help inform diagnostic, prognostic, and therapeutic decision-making for challenging melanoma cases. We present a melanoma concordance regression deep learning model capable of predicting the concordance rate of invasive melanoma or melanoma in-situ from digitized Whole Slide Images (WSIs).  The salient features corresponding to melanoma concordance were learned in a self-supervised manner with the contrastive learning method, SimCLR. We trained a SimCLR feature extractor with 83,356 WSI tiles randomly sampled from 10,895 specimens originating from four distinct pathology labs. We trained a separate melanoma concordance regression model on 990 specimens with available concordance ground truth annotations from three pathology labs and tested the model on 211 specimens. We achieved a Root Mean Squared Error (RMSE) of $0.28\pm0.01$ on the test set. We also investigated the performance of using the predicted concordance rate as a malignancy classifier, and achieved a precision and recall of $0.85\pm0.05$ and $0.61\pm0.06$, respectively, on the test set. These results are an important first step for building an artificial intelligence (AI) system capable of predicting the results of consulting a panel of experts and delivering a score based on the degree to which the experts would agree on a particular diagnosis. Such a system could be used to suggest additional testing or other action such as ordering additional stains or genetic tests. 

\end{abstract}

\section{Introduction}
\label{sec:intro}

More than 5 million diagnoses of skin cancer are made each year in the United States, about 106,000 of which are melanoma of the skin \cite{acs2021stats}. Diagnosis requires microscopic examination of hematoxylin and eosin (H\&E) stained, paraffin wax embedded biopsies of skin lesion specimens on glass slides. These slides can be manually observed under a microscope, or digitally on a Whole Slide Image (WSI) scanned on specialty hardware. The 5-year survival rate of patients with metastatic malignant melanoma is less than 20\% \cite{noone2017cancer}. Melanoma occurs more rarely than several other types of skin cancer, and its diagnosis is challenging, as evidenced by a high discordance rate among pathologists when distinguishing between melanoma and benign melanocytic lesions ($\sim40\%$ discordance rate; e.g. \cite{gerami2014histomorphologic}, \cite{elmore2017pathologists}).  The high discordance rate highlights that greater scrutiny is likely needed to arrive at an accurate melanoma diagnosis, however patients receive diagnoses only from a single dermatopathologist in many instances. This tends to increase the probability of misdiagnosis, where frequent over-diagnosis of melanocytic lesions results in severe costs to a clinical practice and additional costs and distress to patients. \cite{welch2021rapid}. In this scenario, the decision-making of the single expert would be further informed by knowledge of a likely concordance level among a group of multiple experts in a given case under consideration. Additional methods of providing concordance information to healthcare providers could help further inform diagnostic, prognostic, and therapeutic decision-making for challenging melanoma cases. A method capable of predicting the results of consulting a panel of experts and delivering a score based on the degree to which the experts would agree on a particular diagnosis would help reduce melanoma misdiagnosis and subsequently improve patient care. 

The advent of digital pathology has brought the revolution in machine learning and artificial intelligence to bear on a variety of tasks common to pathology labs. Campanella \textit{et al.} \cite{campanella2019clinical} trained a model in a weakly-supervised framework that did not require pixel-level annotations to classify prostate cancer and validated on $\sim10,000$ WSIs sourced from multiple countries.  This represented a considerable advancement towards a system capable of use in clinical practice for prostate cancer. However, some degree of human-in-the-loop curation was performed on their data set, including manual quality control such as post-hoc removal of slides with pen ink from the study. Pantanowitz \textit{et al.} \cite{ibex} used pixel-wise annotations to develop a model trained on $\sim550$ WSIs that distinguishes high-grade from low-grade prostate cancer. In dermatopathology, the model developed in \cite{siva2021} classified skin lesion specimens between six morphology-based groups (including melanoma), was tested on $\sim5099$ WSIs, provided automated quality control to remove WSI patches with pen ink or blur, and also demonstrated that use of confidence thresholding could provide a high accuracy.

The recent application of deep learning to digital pathology has predominately leveraged the use of pre-trained WSI tile representations, usually obtained by using feature extractors pre-trained on the ImageNet \cite{imagenet} data set. The features learned by such pre-training is dominated by features present in natural-scene images, which are not guaranteed to generalize to histopathology images. Such representations can limit the reported performance metrics and affect model robustness. It has been shown in \cite{li2021dual} that self-supervised pre-training on WSIs improved the downstream performance in identifying metastastic breast cancer. 

In this work, we present a deep learning regression model capable of predicting from WSIs the concordance rate of consulting a panel of experts on rendering a case diagnosis of invasive melanoma or melanoma in-situ. The deep learning model learns meaningful feature representations from WSIs through self-supervised pre-training, which are used to learn the concordance rate through weakly-supervised training. 

\section{Methods}

\subsection{Data Collection and Characteristics}
\label{sec:data}
The melanoma concordance regression model was trained and evaluated on 1,412  specimens (consisting of 1,722 WSIs) from three distinct pathology labs. The first lab consists for 611 suspected melanoma specimens from a leading dermatopathology lab in a top academic medical center (Department of Dermatology at University of Florida College of Medicine), denoted as \textit{University of Florida}. The second lab consisted of 605 suspected melanoma specimens distributed across North America, but re-scanned at The Department of Dermatology at University of Florida College of Medicine, denoted as \textit{Florida - External}. The third lab consisted of 319 suspected specimens from Jefferson Dermatopathology Center, Department of Dermatology \& Cutaneous Biology, Thomas Jefferson University denoted as \textit{Jefferson}.  The WSIs consisted exclusively of H\&E-stained, formalin-fixed, paraffin-embedded dermatopathology tissue and were all scanned using a 3DHistech P250 High Capacity Slide Scanner at an objective power of 20X, corresponding to 0.24$\mu$m/pixel. The diagnostic categories present in our data set are summarized in Table \ref{tab:diag_counts}.

\begin{table}

    \centering
    \begin{tabular}{l|r}
    Diagnostic Morphology & Number of Specimens \\
    \hline
    Melanoma In Situ &  607 \\
    Invasive Melanoma &  306 \\
    Mild-to-moderate Dysplastic Nevus    & 209 \\
    Conventional Melanocytic Nevus & 123 \\
    Severe Dysplastic Nevus & 49 \\
    Spitz Nevus with Spindle Cell Morphology & 43 \\
    Spitz Nevus & 36 \\
    Junctional Nevus & 27 \\
    Dermal Nevus & 23 \\
    Blue Nevus & 20 \\
    Halo Nevus & 11 \\
    \hline
    Total & 1412
    \end{tabular}
    \caption{Specimen counts of each of the pathologies in the data set, broken-out into specific diagnostic categories.}
    \label{tab:diag_counts}
\end{table}

The annotations for our data set were provided by at least three board-certified pathologists who reviewed each melanocytic specimen. The first review was the original specimen diagnosis made via glass slide examination under a microscope. At least two and up to four additional dermatopathologists independently reviewed and rendered a diagnosis digitally for each melanocytic specimen. The patient's year of birth and gender were provided with each specimen upon review. Two dermatopathologists from the United States reviewed all 1,412 specimens in our data set and up to two additional dermatopathologists reviewed a subset of our data set. A summary of the number of concordant reviews in this study is given in Table \ref{tab:concordant_reviews}. 

The concordance reviews are converted to a concordance rate by calculating the fraction of dermatopathologists who rendered a diagnosis of melanoma in-situ or invasive melanoma. The concordance rate runs from $0.0$ (No dermatopathologist rendered a melanoma in-situ/invasive melanoma diagnosis) to $1.0$ (all dermatopathologists rendered a melanoma in-situ/invasive melanoma diagnosis). It has been previously noted \cite{mpath} that the concordance rate itself is correlated with the likelihood that a specimen is malignant. The concordant labels present in our data set include $0.0$, $0.25$, $0.33$ , $0.5$, $0.67$ ,$0.75$, and $1.0$. For training, validating, and testing, we divided this data set into three partitions by sampling at random without replacement with 70\% of specimens used for training, and 15\% used for each of validation and testing. 990 specimens were used for training, 211 specimens were used for validation, and 211 specimens were used for testing. 

\begin{table}

    \centering
    \begin{tabular}{c|c}
    Number of Dermatopathologists & Number of Specimens \\
    \hline
    Three  &  687 \\
    Four &  216 \\
    Five & 509 \\
    Total & 1412
    \end{tabular}
    \caption{Number of concordant reviews in our data set.}
    \label{tab:concordant_reviews}
\end{table}

\subsection{Melanoma Concordance Regression Deep Learning Architecture}

\begin{figure}[htb]
    \centering
    \includegraphics[width=\textwidth]{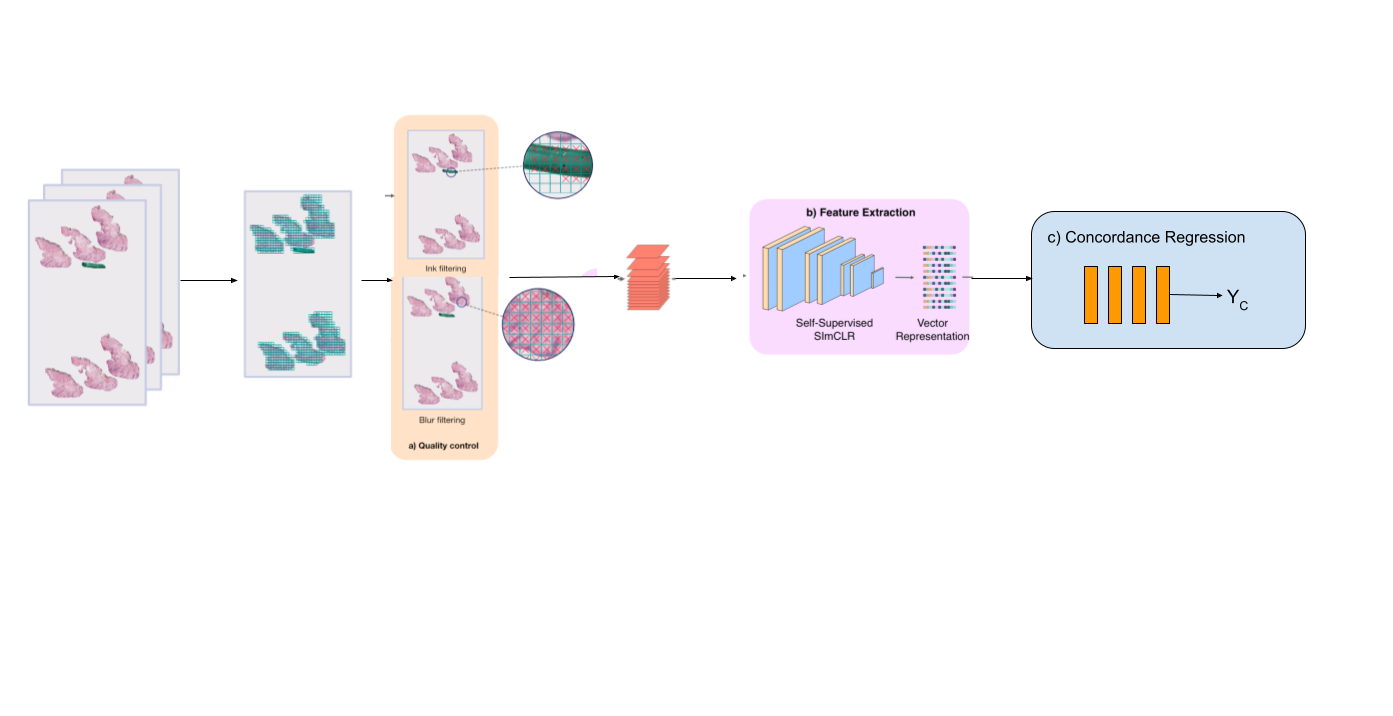}
    \caption{The stages of the Melanoma Concordance Regression pipeline are: Quality Control, Feature Extraction, and Regression. All single specimen WSIs were first passed through the tiling stage, then the quality control stage consisting of ink and blur filtering. The filtered tiles were passed through the feature extraction stage consisting of a self-supervised SimCLR network pre-trained on WSIs with a ResNet50 backbone to obtain embedded vectors. Finally, the vectors were propagated through the regression stage consisting of fully connected layers to obtain a concordance prediction.}
    \label{fig:melcon_pipeline}
\end{figure}

The Melanoma Concordance Regression deep learning pipeline consists of three main components: quality control, feature extraction and concordance regression. A diagram of the pipeline is shown in Figure \ref{fig:melcon_pipeline}.  Each specimen was first segmented into tissue-containing regions through Otsu's method \cite{otsu}, subdivided into 128x128 pixel tiles, and extracted at an objective power of 10X. Each tile was passed through the quality control and feature extraction components of the pipeline.

\subsubsection{Quality Control}

Quality control consisted of ink and blur filtering. Pen ink is common in labs migrating their workload from glass slides to WSIs where the location of possible malignancy was marked. This pen ink represented a biased distractor signal in training that is highly correlated with melanoma. Tiles containing pen ink were identified by a weakly supervised model trained to detect inked slides. These tiles were removed from the training and validation data and before inference on the test set. We also sought to remove areas of the image that were out of focus due to scanning errors by setting a threshold on the variance of the Laplacian over each tile \cite{pertuz2013AnalysisOF}, \cite{pech2000diatom}. 

\subsubsection{Self-supervised Feature Extraction}
\label{sec:simclr}

The next component of the Melanoma Concordance Regression pipeline extracted informative features from the quality controlled tiles. To capture higher-level features in these tiles, we trained a self-supervised feature extractor based on the contrastive learning method proposed in \cite{simclr},\cite{simclrv2} known as SimCLR. SimCLR relies on maximizing agreement in the latent space between representations of two augmented views of the same image. In particular, we maximized agreement between two augmented views of 128x128 pixel tiles in our data set. 

In order to capture as much variety in real-world WSIs as possible, we trained a dermatopathology feature extractor neural network with the SimCLR contrastive learning strategy using skin specimens from four labs and three distinct scanners. The skin specimens originated from both from a sequentially-accessioned workflow from these labs as well as a curated data set to capture the morphological diversity of different skin pathologies. The curated data set included various types of basal and squamous cell carcinomas, benign to moderately atypical melanocytic nevi, atypical melanocytic nevi, melanoma in-situ, and invasive melanoma. We note that the feature extractor training set consisted of wider variety of skin pathologies than the concordance regression model, which was trained and tested only on the skin pathologies outlined in Table \ref{tab:diag_counts}. We included WSIs from the University of Florida and Jefferson that were scanned with 3D Histech P250 scanners. We also included WSIs from another top medical center, the Department of Pathology and Laboratory Medicine at Cedars-Sinai Medical Center, which were scanned with a Ventana DP 200 scanner. We finally included WSIs from an undisclosed partner lab in western Europe that were scanned with a Hamamatsu NanoZoomer XR scanner. We note that WSIs from Cedars-Sinai and the undisclosed partner lab were only used for training the feature extractor and not the concordance regression model, as concordance review annotations were not available for these labs. 

 We randomly sampled 26,209 tiles from Florida and 57,147 tiles from the remaining three labs for a total of 83,356 tiles randomly sampled from 10,895 specimens for use during training of the feature extraction network. Each tile was sampled from WSIs extracted at an objective power of 10x. We set the temperature hyperparameter, ($\tau$) to $\tau = 0.1$, batch size $= 128$, and the learning rate $=0.001$ during training. (We note that the feature extractor is trained separately from the Melanoma Concordance Regression model described in section \ref{sec:mil}.) We randomly divided $80\%$ of the tiles for training, and $20\%$ for validation.  We used the ResNet50 \cite{resnet50} backbone for training, and the Normalized Temperature-scaled Cross Entropy (NT-Xent) loss function as in \cite{simclr}. We followed the same augmentation strategies in \cite{simclrv2} applied to the tiles. We used a temperature hyperparemter value of $0.1$. We achieved a minimum NT-Xent loss value on the validation set of $2.6$. As a point of comparison, we note that the minimum validatation NT-Xent loss when training with the ImageNet data set with a ResNet50 backbone is $4.4$ \cite{pytorchlightning}. However, given that tiles from WSIs are vastly different from landscape images in ImageNet, this may not be meaningful. Once the feature extractor was trained, it was deployed to the Melanoma Concordance Regression pipeline in order to embed each tile from the melanoma consensus regression data set (consisting of 1,722 WSIs) into a latent space of 2048-channel vectors.

\subsubsection{Melanoma Concordance Regression}
\label{sec:mil}
The melanoma concordance regression model predicted a value representing the fraction of dermatopathologists that is concordant with a diagnosis of Melanoma In-Situ or Invasive Melanoma.  The model consisted of four fully-connected layers (two layers of 1024 channels each, followed by two of 512 channels each). Each neuron in these four layers was ReLU activated. The model was trained under a weakly-supervised multiple-instance learning (MIL) paradigm. Each embedded tile propagated through the feature extractor described in section \ref{sec:simclr} was treated as an instance of a bag containing all quality-assured tiles of a specimen. Embedded tiles were aggregated using sigmoid-activated attention heads \cite{ilse2018attention}. The final layer after the attention head was a linear layer that takes 512 channels as input and outputs a concordance rate prediction. To help prevent over-fitting, the training data set consisted of augmented versions (inspired by Tellez \textit{et al.}\cite{tellez}) of the tiles. Augmentations were generated with the following augmentation strategies: random variations in brightness, hue, contrast, saturation, (up to a maximum of 15\%),  Gaussian noise with 0.001 variance, and random 90 degree image rotations.  We trained the melanoma concordance regression model with the Root mean squared error (RMSE) loss function. We regularized model training by using the dropout method \cite{dropout} with $20\%$ probability.

\section{Results}

\begin{figure}[htb]
    \centering
    \includegraphics[width=\textwidth]{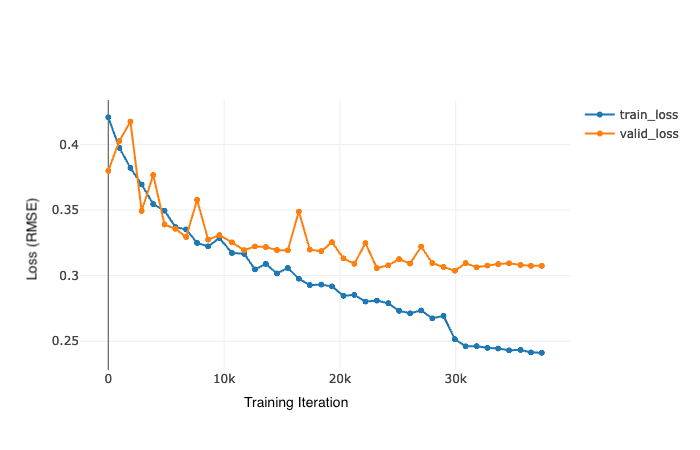}
    \caption{Melanoma Concordance Regression Training and Validation Loss Curves. The minimum validation RMSE loss was found to be $0.30$. The RMSE loss on the test set was found to be $0.28\pm0.01$. }
    \label{fig:loss_curves}
\end{figure}

\begin{table}
    \centering
    \begin{tabular}{c|c|c|c|c}
    Metric & All Sites & University of Florida & Florida - External & Jefferson \\
    \hline
    RMSE & $0.28\pm0.01$ & $0.27\pm0.02$ & $0.26\pm0.02$ & $0.3\pm0.03$ \\
    $R^{2}$ & $0.51\pm0.05$ & $0.49\pm0.09$ & $0.52 \pm 0.08$ & $0.36\pm0.13$\\
    \hline
    \end{tabular}
    \caption{Regression metrics across the individual laboratory sites in our test data set. Errors are 90\% confidence intervals derived through bootstrapping.}
    \label{tab:reg_metrics_lab}
\end{table}

\begin{figure}[htb]
    \centering
    \includegraphics[width=\textwidth]{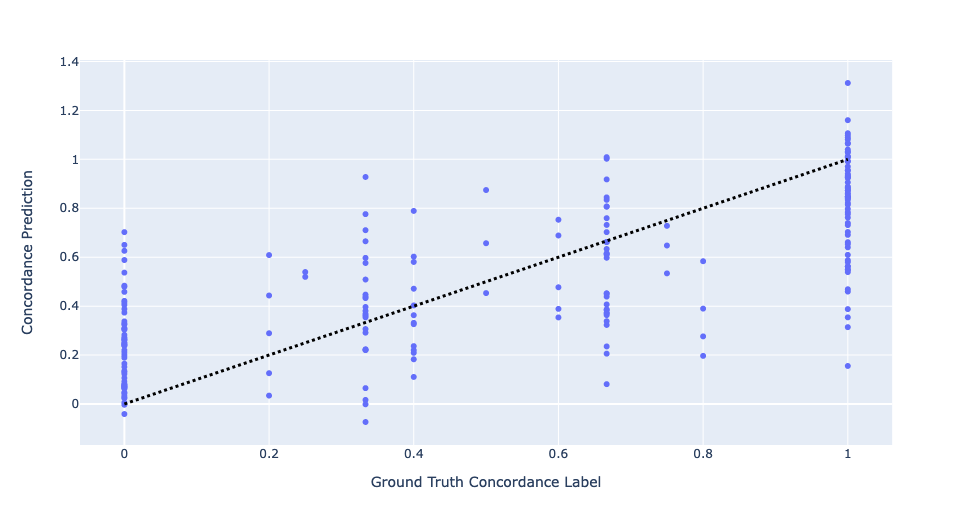}
    \caption{The correlation of melanoma concordance predictions against the ground truth concordance label.}
    \label{fig:preds_vs_labels_all}
\end{figure}

To demonstrate the performance of the Melanoma Concordance Regression model, we first show the training and validation loss curves in Figure \ref{fig:loss_curves}. The minimum validation loss was found to be $0.30$. We then calculated the RMSE and $R^{2}$ on the test set both across laboratory sites and for individual laboratory sites with 90$\%$ confidence intervals derived from bootstrapping with replacement. The RMSE across laboratory sites on the test set was calculated to be $0.28\pm0.01$ and the $R^{2}$ was found to be $0.51\pm0.05$. The lab-specific regression performance of the model on the test set is summarized in Table \ref{tab:reg_metrics_lab}. We note that the RMSE performance is consistent across sites and are within the error bars derived from bootstrapping within replacement. The correlation between the melanoma concordance predictions and the ground truth labels is shown in Figure \ref{fig:preds_vs_labels_all}. 

\begin{figure}[htb]
    \centering
    \includegraphics[width=0.7\textwidth]{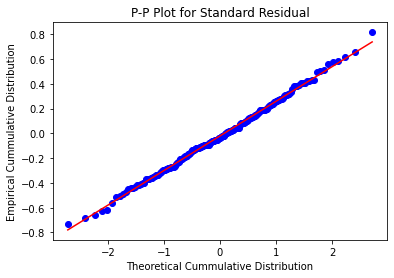}
    \caption{A Probability-Probability plot is used as a goodness-of-fit in order to compare the cumulative distribution function of the standardized residuals of the melanoma concordance regression model to a Gaussian distribution. The P-P plot demonstrates the standardized residuals of melanoma concordance regression are normally distributed. The residuals were found to be consistent with a Normal distribution by performing the Shapiro-Wilk test on the standardized residuals. The p-value of the test was found to be $0.97$.}
    \label{fig:pp_plot}
\end{figure}

We next assessed the goodness-of-fit of the regression model by calculating the standardized residuals:

\begin{align}
  e_{i} & = \frac{y_{pred,i} - y_{true,i}}{\sigma} 
\end{align}

where $i$ is the ith data point, and $\sigma$ is the standard deviation of the residuals. The P-P plot comparing the cumulative distribution functions of the standardized residuals to a Normal distribution is shown in Figure \ref{fig:pp_plot}. The standardized residuals were found to be consistent with a Normal distribution by performing the Shaprio-Wilk test \cite{shapiro}, where the p-value was found to be $0.97$.

\subsection{Malignant Classification}
\label{sec:res_mag}

\begin{table}[htb]
    \centering
    \begin{tabular}{c|c|c|c|c}
    Metric & All Sites & University of Florida & Florida - External & Jefferson \\
    \hline
    AUC  &  $0.89\pm0.02$  & $0.86\pm0.04$ & $0.9\pm0.04$ & $0.85\pm0.06$ \\
    Precision &  $0.86\pm0.05$ & $0.89\pm0.08$ & $1.0\pm 0.0$ & $0.7 \pm 0.15$\\
    Recall &$0.61\pm0.06$ & $0.65\pm0.09$ & $0.5\pm0.1$ &$0.74 \pm 0.12$ \\
    Specificity& $0.97\pm0.02$ & $0.94\pm0.04$ & $1.0\pm0.0$ & $0.9\pm0.05$
    \end{tabular}
    \caption{Classification metrics when using the predicted concordance value as a malignant classifier across the individual laboratory sites in our test data set. Errors are 90\% confidence intervals derived through bootstrapping with replacement.}
    \label{tab:mag_metrics}
\end{table}

\begin{figure}[htb]
    \centering
   \centering
    \includegraphics[width=0.45\columnwidth]{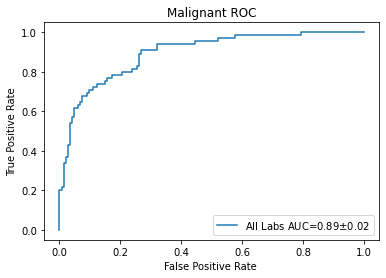}
    \includegraphics[width=0.45\columnwidth]{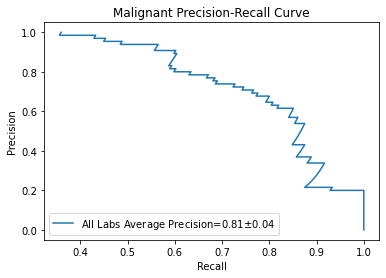}
    \includegraphics[width=0.45\columnwidth]{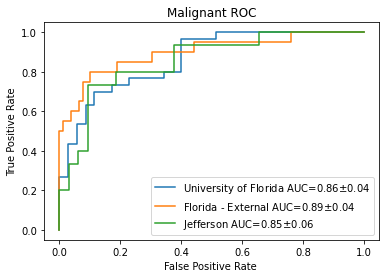}
    \includegraphics[width=0.45\columnwidth]{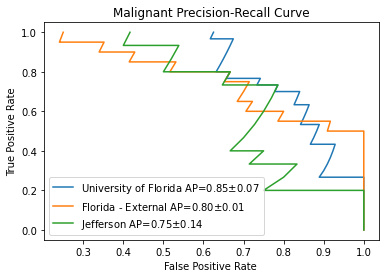}
    \caption{Receiver Operating Characteristic (ROC) and Precision-Recall curves for using the melanoma concordance prediction as a malignant classifier on the test data set. The top row shows the curves across laboratory sites, and the bottom row shows the curves for individual sites. The AUC value for malignant classification was found to be $0.89\pm0.02$ and the average precision (AP) was found to be $0.81\pm0.04$. The bottom row demonstrates the curves for individual laboratory sites. }
    \label{fig:auc_pr}
\end{figure}

As mentioned in section \ref{sec:data}, increased inter-pathologist agreement on melanoma correlates with malignancy. We therefore investigated the performance of using the predicted concordance rate of the melanoma concordance model as a binary classifier to classify malignancy. We derived malignancy binary ground truth labels by defining a threshold value to binarize the ground truth concordance label, where a specimen with an observed concordance rate above this threshold value received a label of malignant, and below this threshold value received a label of not malignant. We performed a grid search on possible ground truth thresholds, and chose the threshold that maximized the Area Underneath the Receiver Operating Characteristic (ROC) Curve (AUC). We also found the same ground truth threshold  maximized the average precision of the precision-recall curve. We found that a threshold of $0.85$ on the ground truth concordance rate to label malignancy maximizes both AUC and average precision, and yielded an AUC value of $0.89\pm0.02$ and an average precision of $0.81\pm0.04$.  The classification metrics with this threshold are shown both across laboratory sites and individual laboratory sites in Table \ref{tab:mag_metrics}. The ROC and precision-recall curves of using the melanoma concordance prediction as a malignant classifier is shown in Figure \ref{fig:auc_pr} both across laboratory sites as well as individual laboratory sites. We note that the classification performance are within the error bars derived from bootstrapping with replacement across sites, with the exception of the low false positive rate at Florida - External, which exhibits very high precision. 

\subsection{Ablation Studies}

\subsubsection{Binary Classifier Ablation Study}

\begin{table}[htb]
\centering
    \begin{tabular}{c|c|c|c}
    Metric & Melanoma Concordance Regression & Binary Classifier \\
    \hline
    AUC  &  $0.86\pm0.04$  &   $0.67\pm0.03$ \\
    Precision &  $0.89\pm0.08$ &  $0.57\pm0.07$  \\
    Recall & $0.62\pm0.08$ &  $0.51\pm0.07$ \\
    Specificity& $0.97\pm0.02$ &  $0.75\pm0.04$ 
    \end{tabular}
    \caption{Results of the binary classifier ablation experiment. A binary malignancy classifier was trained with the cross entropy loss function on data from the University of Florida, and the results are compared to using the predicted concordance rate from the regression model to classify malignancy.}
 \label{tab:mag_metrics_binary}
\end{table}
We compared the performance of using the concordance rate as a malignancy classifier with a model trained to perform a binary classification task. In particular, we configured the final linear layer of the deep learning model to predict one of two classes: malignant or not malignant.  We subsequently trained a binary classification model with the cross entropy loss function. We used a threshold of $0.85$ on the ground truth to annotate malignancy derived from Section \ref{sec:res_mag}.  We performed this ablation study specifically on data from the University of Florida, and the results are shown in Table \ref{tab:mag_metrics_binary}. It can be seen that using the concordance rate to classify malignancy yields better performance than training a dedicated binary classifier. 

\subsubsection{Feature Extractor Ablation Study}

\begin{figure}[htb]
   \centering
    \includegraphics[width=1.0\columnwidth]{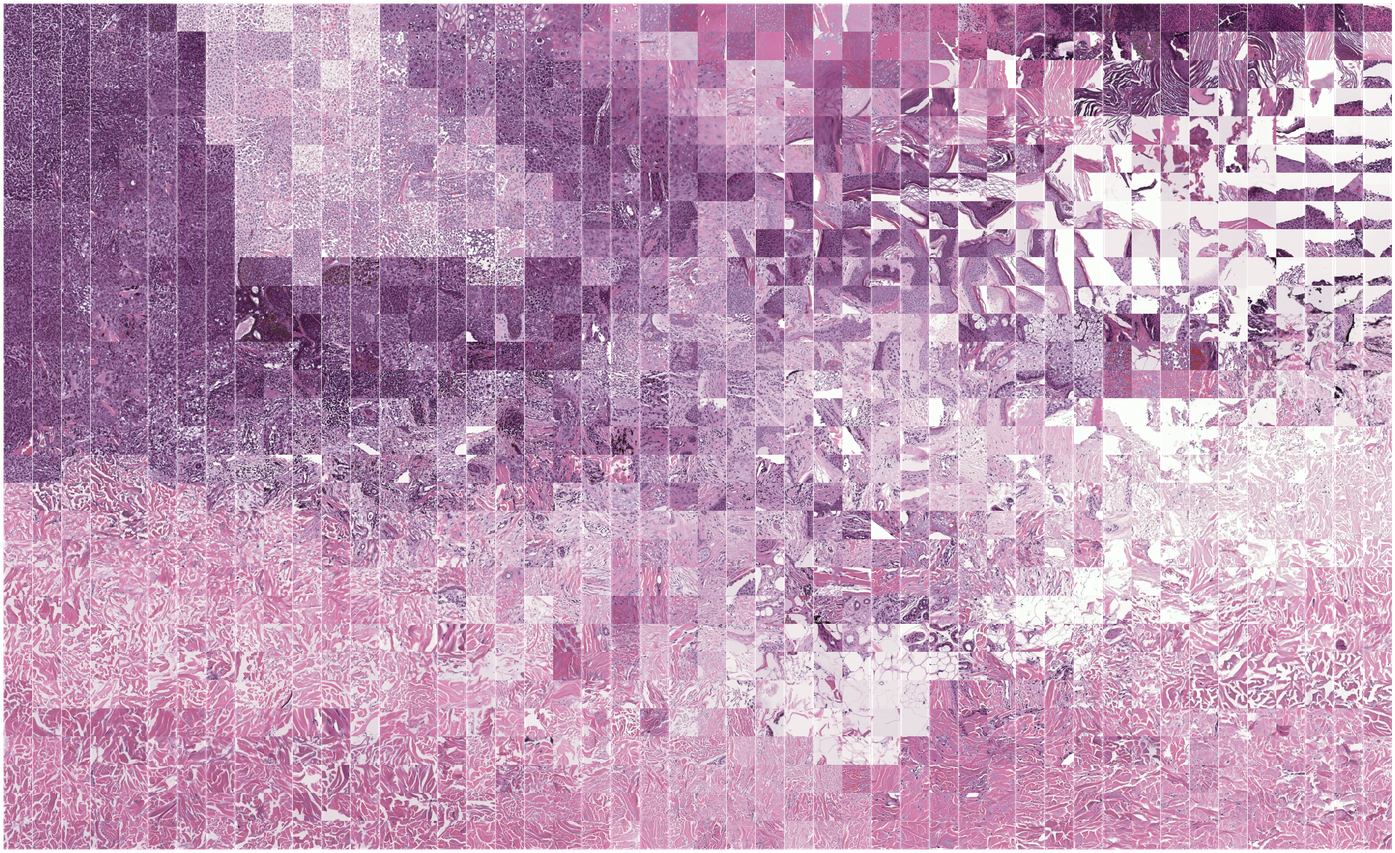}
    \caption{Grid of SimCLR feature vectors arranged by their embedded vector values. First, SimCLR feature vectors of 1,440 selected tiles were projected onto two dimensions with the UMAP algorithm. We next arranged the WSI tiles on a grid where the two-dimensional UMAP coordinates are mapped to the nearest grid coordinate, which arranged the tiles onto an evenly spaced plane.}
    \label{fig:simclr_grid}
\end{figure}

We mentioned in Section \ref{sec:intro} that features learned by pre-training on ImageNet are not guaranteed to generalize to histopathology image and could limit performance metrics and model robustness. To test this claim, we first visually examined SimCLR feature vectors to investigate visual coherence with morphological features. In particular, we projected the SimCLR feature vectors of 1,440 selected tiles to two dimensions with the Uniform Manifold Approximation and Projection (UMAP \cite{umap}) algorithm. We then arranged the tiles on a grid where the two-dimensional UMAP coordinates are mapped to the nearest grid coordinate, which arranged the tiles onto an evenly spaced plane. The arranged tiles are shown in Figure \ref{fig:simclr_grid}. It can be seen that the learned feature vectors displays a coherence with respect to morphological features. 

\begin{table}[htb]
\centering
    \begin{tabular}{c|c|c|c}
    Metric & SimCLR trained with WSIs &  ResNet50 trained with ImageNet \\
    \hline
    RMSE &  $0.27\pm0.02$ &   $0.31\pm0.02$ \\
    $R^{2}$ &  $0.49\pm0.09$ &   $0.15\pm0.15$ \\
    \hline
    AUC  &  $0.86\pm0.04$  &  $0.75\pm0.06$\\
    Precision &  $0.89\pm0.08$ &  $0.76\pm0.07$ \\
    Recall & $0.62\pm0.08$ &  $0.45\pm0.08$ \\
    Specificity& $0.97\pm0.02$ & $0.71\pm0.08$
    \end{tabular}
    \caption{Feature Extractor ablation study results from the University of Florida. feature vectors derived from SimCLR with a ResNet50 model backbone trained on Whole Slide Images were compared to feature vectors derived from a pre-trained ResNet50 model on the ImageNet data set. Concordance regression metrics and classification metrics are shown. Classification metrics were derived by utilizing the predicted concordance value as a malignant classifier.}
    \label{tab:mag_metrics_imagenet}
\end{table}

We next performed an ablation experiment to quantify the impact that our self-supervised feature extractor has on model performance. In particular, we ablated the SimCLR embedder by propagating the WSI tiles specifically from the University of Florida through a ResNet50 \cite{resnet50} pre-trained on the ImageNet \cite{imagenet} data set to embed each input tile into 2048 channel vectors. The embedded vectors were used to train a subsequent melanoma concordance regression model. The resulting model performance for this feature ablation experiment is summarized in Table \ref{tab:mag_metrics_imagenet} for both the raw regression metrics as well as the malignant classification task. (The same $0.85$ threshold on the ground truth concordance rate defined in \ref{sec:res_mag} was used to define malignancy) It can be seen that SimCLR results in higher performance in both the concordance regression and malignant classification tasks. In particular, there is a $14\%$ improvement in RMSE in using the SimCLR embedder over imagenet for the regression task, and a $27.5\%$ improvement in recall for the malignant classification task.

\section{Conclusions}

We presented in this paper a melanoma concordance regression model that was trained across three laboratory sites that demonstrates regression performance of $0.28\pm0.01$ on the test set across the three laboratory sites in our data set. The performance of the melanoma concordance regression model was limited by the number of dermatopathologists performing concordant reviews on our data set. From Table \ref{tab:concordant_reviews}, the average number of dermatopathologists reviewing a case is $3.7$ pathologists, resulting in a concordance rate resolution of $0.27$. The melanoma concordance regression model that we built is at the limit of this resolution.  The regression performance could be further improved by collecting more concordance reviews from dermatopathologists. We maximized the melanoma concordance regression performance by learning informative features through self-supervised training with SimCLR. We also demonstrated that the predicted melanoma concordance rate can be used as a malignant classifier, as the concordance rate is correlated with the likelihood of malignancy. We adjusted the threshold of malignancy to maximize both AUC and average precision, where we found the AUC value to be $0.89\pm0.02$ and the average precision to be $0.81\pm0.04$.

These results are an important first step for building an AI system capable of predicting the results of consulting a panel of experts and delivering a score based on the degree to which the experts would agree on a particular diagnosis or opinion. Upon further improvement, a concordance score reliably representing a panel of experts can additionally be used as a diagnostic assist to suggest additional testing or other action, for example the ordering of additional stains or a genetic test. Additionally, dermatopathologists utilizing a concordance score as a diagnostic assist can empower them to speed up malignancy diagnosis, increase confidence in case sign out, and lower the misdiagnosis rate, thereby increasing sensitivity and improving patient care. The possibility also exists that a concordance score could be fed into a combined-test that incorporates the results of a multi-gene assay (e.g., Castle MelanomaDx \cite{mypath}) in order to enhance the performance of the test. Finally, such a concordance AI system can be extended to any pathology that exhibits a high discordance rate, such as breast cancer staging \cite{brcaconcord} and Gleason grading of prostate cancer \cite{gleasoncon}.

\clearpage
%
%
\bibliographystyle{splncs04}
\bibliography{bibliography.bib}
\end{document}